# Understanding Meanings in Multilingual Customer Feedback

**Chao-Hong Liu**
ADAPT Centre, Ireland
Dublin City University
`chaohong.liu`
`@adaptcentre.ie`

**Declan Groves**
Microsoft Ireland
Leopardstown, Dublin 18
`degroves`
`@microsoft.com`

**Akira Hayakawa**[1]**, Alberto Poncelas**[2] **and Qun Liu**[2]
ADAPT Centre, Ireland
[1] Trinity College Dublin
[2] Dublin City University
`{akira.hayakawa,`
`alberto.poncelas,`
`qun.liu}@adaptcentre.ie`

## Abstract

Understanding and being able to react to customer feedback is the most fundamental task in providing good customer service. However, there are two major obstacles for international companies to automatically detect the meaning of customer feedback in a global multilingual environment. Firstly, there is no widely acknowledged categorisation (classes) of meaning for customer feedback. Secondly, the applicability of one meaning categorisation, if it exists, to customer feedback in multiple languages is questionable. In this paper, we extracted representative real-world samples of customer feedback from Microsoft Office customers in multiple languages, English, Spanish and Japanese, and concluded a five-class categorisation (comment, request, bug, complaint and meaningless) for meaning classification that could be used across languages in the realm of customer feedback analysis.

## 1 Introduction

In this paper we discuss the results of an ADAPT-Microsoft joint research project which aims to assess the performance of the internal tools of Microsoft on multilingual customer feedback analysis. The current approach to multilingual customer feedback analysis is to translate non-English feedback into English using machine translation (MT) systems and use English-based tools for the analysis. Due to the variability of MT quality, in particular when translating user-generated content, a reduction in the ability of such tools to accurately analyse customer feedback in non-English languages is to be expected. To test this assumption, two languages, Spanish and Japanese, were considered in this assessment. In summary, the tools perform well if MT quality is good in general as exemplified by Spanish customer feedback. The true positive rate is about 20% lower in Japanese feedback where the lower MT quality might be one of the causes of the comparative underperformance.

The results suggest that building native tools might be necessary for languages where MT quality is not satisfactory. To do this, a corpus of annotated customer feedback for each language should be prepared, which raises two questions; 1) what categorisation of customer feedback should be used as annotation scheme, and 2) does this categorisation apply to multiple languages? To answer these questions and to improve the ability to understand international customer feedback, we summarised directly from free-form meaning annotations on both Spanish and Japanese feedback and concluded a five-class categorisation (comment, request, bug, complaint, and meaningless) which might be used across languages, including English, for future customer feedback analysis.

Sentiment analysis itself has commonly been used in customer feedback analysis as part of meaning-based analysis in Microsoft Office and in categorisation approaches within other companies and organisations (Salameh et al., Afli et al.) In this paper however, we separate sentiment from what was expressed, the meanings or intentions, in customer feedback. Customer feedback analysis nowadays has become an industry in its own right; there are dozens of notable internet companies (which we refer to as 'app companies') who are performing customer feedback analysis for other, often much larger, companies.

The business model for these app companies is to acquire customer feedback data from their clients and to perform analysis using their internal tools and provide reports to their clients periodically (Freshdesk, Burns). However, most app companies not only treat the contents of these reports as confidential material, which is understandable, but also regard things such as the set of categories they used for grouping customer feedback as business secrets.

To the best of our knowledge, there are three different openly available categorisations from these app companies. The first is the most commonly-used categorisation which could be found in many websites, i.e. the five-class Excellent-Good-Average-Fair-Poor and its variants (Yin et al., SurveyMonkey). The second one is a combined categorisation of sentiment and responsiveness, i.e. a five-class Positive-Neutral-Negative-Answered-Unanswered, used by an app company (Freshdesk) The third one is used by another app company called Sift and the categorisation is a seven-class Refund-Complaint-Pricing-Tech Support-Store Locator-Feedback-Warranty Info (Keatext). There are certainly many other possible categorisations for customer feedback analysis, however, most of them are not publicly available (Equiniti, UseResponse, Inmoment).

In this paper, we try to answer the question if there is a suitable categorisation of meanings for customer feedback which could be used in multiple languages. In Section 2, we give brief description on how we acquire the corpora of multilingual customer feedback. The observations of meanings in Japanese and Spanish customer feedback are presented in Section 3 and 4. Section 5 details the summarised five-class categorisation which we propose to use for multilingual customer feedback analysis as a common annotation scheme in the future. Finally, the conclusions are given in Section 6.

## 2 Preparation of Customer Feedback Corpora

Microsoft Office collects customer feedback via several different channels, which is aggregated and analysed via the internal Office Customer Voice (OCV) system. OCV gathers submitted user data across 70+ languages, and implements classification and on-demand clustering, using semi-supervised techniques, together with a rich web UI to facilitate further analysis and reporting by product owners. These product owners can carry out trend analysis to identify issues that require specific attention or feature requests from users. Due to the large quantities of customer feedback received on a regular basis, manual processing is not feasible. Therefore, the ability to quickly identify meaning is key to help establish the actionability and importance of customer feedback.

The system implements supervised logistic regression to create two multi-class models; one for area and one for issue (Bentley & Batrya, 2016). Additionally, inferences are provided via Sys-Sieve, an internal inference engine (Potharaju, Jain & Nita-Rotaru, 2013) and two-class sentiment analysis via Azure ML's sentiment classifier.[1] As mentioned previously, the system operates only on English language feedback. For non-English feedback, which constitutes on average 58% of total monthly feedback in Microsoft Office, MT is used as a pre-processing technology provided via Microsoft's general-domain Translator APIs.[2]

To illustrate the classification process, Bentley & Batrya (2016) provide the example of the verbatim feedback "I am having trouble saving an Excel spreadsheet with charts in it." OCV assigns it the possible issue types of "Excel\Charts" and "Excel\Save", but not "Excel\Print", whereas Sys-Sieve may infer that it relates to "Problem" and Azure ML's sentiment classifier determines that it has a higher probability of carrying "Negative" sentiment, than "Positive". This type of coarse-grained classification is typically sufficient for a product owner to identify emerging trends and issues that require subsequent triage.

### 2.1 Data Selection

In terms of content, for this study we made the decision to focus on feedback received via the "sent-a-smile" feature which is the prime source of in-application ("in-app") feedback for Microsoft Office products. This allows a user to provide feedback directly from within the application, including the provision of verbatim textual feedback and screenshot information if the user so wishes to do so.

---

[1] https://studio.azureml.net

[2] https://www.microsoft.com/en-us/translator/translatorapi.aspx

We chose two target languages to focus on: Spanish and Japanese. Based on previous internal qualitative evaluations, they represent contrasting languages with respect to the expected quality of MT; Spanish content typically preforms very well, whereas Japanese is a more difficult language for MT.

Table I: MT Quality Scale

| Score | Adequacy | Fluency |
|---|---|---|
| 4 | All meaning of the source correctly expressed in the translation | Completely fluent. Good word choice & structure. No editing required. |
| 3 | Most of the source expressed in the translation | Almost fluent. Few errors which don't impact the overall meaning. |
| 2 | Little of the source expressed in the translation | Not very fluent. About half the translation contains errors & requires editing. |
| 1 | No source meaning expressed in the translation | Incomprehensible. Needs to be translated from scratch. |

Taking an initial 12-month snapshot of 4,254 Japanese and 28,352 Spanish pieces of in-app feedback for the same product, we randomly sampled 2,000 items for each language (i.e. 4,000 pieces in total), ensuring that the items selected had been assigned a label by OCV's automatic classifiers and by the inference engine (i.e. we excluded any manually labelled items or items that for any reason were not assigned a label). We subsequently had the quality of the MT'd verbatim feedback judged by human evaluators who assigned both fluency and adequacy scores on a scale of 1-4 (cf. Table I). We did not carry out any filtering on source quality, but it is acknowledged that typographical errors in the user input, slang, idiomatic expressions and abbreviations can all have considerable impact on the comprehensibility of the translation, thus why the MT quality for this domain may often be lower than expected.

Table 2 provides the results for human MT quality judgements for Spanish and Japanese feedback. We performed some initial analyses to measure the impact of MT quality on classification accuracy (for area, issue and sentiment) and found overall, on a per-language basis, that although improving MT quality does result in improvements in area, issue and sentiment classification accuracy (an overall improvement of approx. 10% was observed, on average, with Japanese benefiting more than Spanish), the impact was not significant to warrant the exclusion of any of the data from further analyses i.e. we can get useful classification even with less than perfect MT output.

Table 2: MT Quality for Customer Feedback

| Language | Fluency | Adequacy | Mean |
|---|---|---|---|
| Spanish | 2.89 | 3.16 | 3.03 |
| Japanese | 2.35 | 2.46 | 2.41 |

The feedback items were also manually labelled by human annotators for area and inference type. OCV will typically automatically assign a large number of potential area classes to each item, together with a probability score derived from the classifier indicating the likelihood that the verbatim text belongs to that class. To ensure there were no data sparsity issues, we mapped the large initial set of area classes to a smaller set of 16 (the human annotators were requested to use this smaller set). Sentiment consistent of three classes (positive, negative and neutral) and there were 7 possible inference types (e.g. "problem", "delighter", "suggestion").

## 3 Analysis of Meanings in Japanese Customer Feedback

In this section, we use meanings instead of "Issues" for discussion purposes. A native speaker of Japanese was asked to annotate the meaning of each customer feedback (item); no pre-defined taxonomy was given and the native speaker was instructed to add or modify the "meanings" if they found it is necessary or more appropriate to do so. It was recommended to the annotators for both Spanish and Japanese to aim for a small set of master labels in order to mitigate the possibility of data sparseness in future analyses. The resulting taxonomy of the meanings is as follows:

1. Opinion/Comment (662)
2. Complaint (568)
3. Request (274)
4. NA/No meaning (32)
5. Appreciation (3)
6. Apology (1)
7. Sarcasm (1)

The large majority of items were annotated with multiple labels as feedback often reflects multiple meanings. It is interesting to see that feedback items intended to give ideas (opinion/comment) and to request improvements to the software comprise approximately two thirds of the items (936 mentions), while complaints relate to approximately only a third (568 mentions).

According to the native speaker, there are two clearly distinct genres in the Japanese feedback. One is from casual users, or consumers, of Microsoft Office software and the other, task critical users, typically representing enterprise customers. Feedback of the second genre tends to be polite and gives useful information that could be used to improve the software. This could be part of the reason why these items (opinion/comment and request) comprise the bulk of the feedback.

We also looked into detail for some of the major semantic classes. Here are the fine-grained semantic sub-classes for "Request" and "Complaint":

1. Request:
   a. Add feature
   b. User's guide
   c. B2C communication
   d. Feature change
   e. Standardisation
   f. Compatibility/Hardware compatibility
   g. Solution to reliability problem
   h. Improvement
2. Complaint:
   a. Add feature
   b. Overall performance/Software performance
   c. Bug report
   d. UI/UI design
   e. Feature change/Feature setting
   f. B2C communication
   g. Customer service
   h. Standardisation
   i. Improvement
   j. Product concept
   k. Printout display
   l. Usability
   m. MS server (generalised as software interoperability problem)
   n. Wrong usage of Japanese (generalised as language usage problem)

This 7-class taxonomy of meanings and its fine-grained categorisation are summarised directly from Japanese customer feedback sentences in the Japanese corpus. Although the corpus is mainly for Microsoft Office products, we contend they are general enough for customer feedback analysis for other software products.

### 3.1 Criticality in Japanese Customer Feedback

We also observed a new linguistic concept called "criticality" for the annotation of customer feedback, which applies to both "meanings" and "sentiments." This criticality concept indicates if the customer sees her/his problem as critical to her/his task on hand and requires addressing, regardless if it has been addressed or not. Negative feedback in terms of sentiment might not be of "critical" importance in some cases. For example, a Japanese item which in English translates as "I need a simple manual on how it is used." was annotated as negative sentiment with "minor" criticality.

There are only 40 items annotated as critical in meanings. Most items are not annotated with criticality values and it seems this could be a good indicator to identify which items are of interest to customer service.

1. Critical (40)
2. Medium (87)
3. Minor (72)
4. N/A (648)

The "N/A" (not applicable) refers to those items where criticality is not expressed in the item. For example, a Japanese item with the English translation "Easy to use."

It is interesting to see some critical items and their English translations.

1. Item 1742: "If there was a 'Select File Format and Copy' feature it would be perfect."
2. Item 1904: "Problem solved. For months it was so slow that it cannot do anything. It is great that this is solved in the last update."
3. Item 1977: "All right. (I am) satisfied. Editing and browsing are now running smoothly."

Item 1742 is annotated as "request" (to add a feature) while items 1904 and 1977 are annotated as "appreciation" (on bug fixing) in meanings. The contents showed that the users are either eager to add a feature that would be useful or that they are very satisfied with the improvements of software.

## 4 Analysis of Meanings in Spanish Customer Feedback

In the annotation of Spanish customer feedback, the taxonomy used by OCV internally was not exposed to the native Spanish speaker, either. The native speaker was free to annotate the meaning of each item as appropriate.

The native Spanish annotator noted that users in Spanish-speaking countries often use sarcasm or humour in their responses, and the feedback frequently reflects their tendency towards freedom of expression and frankness. In addition, not knowing *a priori* the cultural background of the users who have provided the feedback makes evaluating the politeness of sentences more complex as different Spanish-speaking regions tend to express themselves in different ways. An example of this is how a speaker refers to other people within the context of feedback: the sentence "son los mejores" (translated as "you are the best") would be considered as neutral tone in Latin American countries, while in Spain it would be polite ("sois los mejores" would be more typical of how someone from Spain would express this in a more casual way).

In our opinion, Spanish feedback lends itself well to the detection and identification of problems when seeking frank opinions from customers after the launch of a new feature or software product. There are 2,051 items in the Spanish corpus; two items are written in Catalan. The resulting taxonomy of meanings is as follows.

1. Congratulate (1243)
2. Request (420)
3. Bug (267)
4. Usability (61)
5. Complaint (28)
6. Nonsense (19)
7. Sarcasm (8)
8. Meaningless (6)

We first noticed the high proportion of "Congratulate" and saw it could be a regional and cultural phenomenon in Spanish-speaking countries. Examples of this feedback include "everything is very useful thank you" and "excellent application". The Spanish feedback is also notable for the customers' short responses, e.g. "I like it", "Good" and "Simple".

There are not many "complaint" types in Spanish feedback when compared to Japanese feedback. "Usability" and "Bug" are the native speaker's own labels, while in Japanese annotation, these two categories are classified as part of "complaint."

The native speaker also distinguished the concepts of "Nonsense" and "Meaningless". In "Meaningless", users expressed messages indicating that they need more time to give proper feedback. An example of this is "I just start using it. I will give my opinions once tried using it for one month". In the "Nonsense", users are inputting texts that are not relevant to customer feedback, e.g. "Best regards" and "Bad don't let me go".

## 5 Common Categorisation of Meanings for Customer Feedback

We summarised the categorisations in Table II for comparison purposes. It seems that despite the cultural differences, meanings can be generalised for both Spanish and Japanese customer feedback, which is comprised of the five classes as follows.

1. Comment (including Congratulate, Apology and Sarcasm)
2. Request (e.g. a new feature or improvement of existing features)
3. Bug (Reporting)
4. Complaint (including Usability)
5. Meaningless (in the contexts of customer feedback)

Table II: Summarised Meaning Categorisation for Customer Feedback

| Common Categorisation | Native Spanish Speaker Categorisation | Native Japanese Speaker Categorisation |
|---|---|---|
| Comment | Congratulate (1243) Usability (61) Sarcasm (8) | Opinion / Comment (662) Appreciation (3) Apology (1) Sarcasm (1) |
| Request | Request (420) | Request (274) |
| Bug | Bug (267) | Bug report (185) |
| Complaint | Complaint (28) | Complaint (383) |
| Meaningless | Nonsense (19) Meaningless (6) | NA / No meaning (32) |

It should be noted, that the five classes above are not necessarily exclusive. For example, a piece of feedback might be both a bug and complaint at the same time. Secondly, the sense of comments is constrained considering other classes. For example, a 'negative comment' will be annotated as a 'complaint' rather than a comment.

Despite the five-class mapping suggested in this paper, it should be noted that in certain circumstances a finer-grained language-specific categorisation might still be of interest. For future work, we plan on investigating whether a larger number of language-specific finer-grained categorisation sets could be combined and generalised to adequately represent multiple languages.

## 6 Conclusions

In this paper, we addressed the problem of understanding the meanings of multilingual customer feedback. Real-world customer feedback from Microsoft Office customers are collected and analysed in three languages. Customer feedback in Spanish and Japanese are annotated by native speakers with the meanings they see fit for the sentences in each feedback text, without any pre-defined categorisation. A five-class categorisation (i.e. comment, request, bug, complaint and meaningless) are summarised from the free-form meaning annotation, which we propose to use as a fundamental annotation scheme for meaning classification for multilingual customer feedback analysis.

For future work, we would like to train a classifier using the suggested annotation scheme and compare the performance of the new classifier against the existing OCV classification. Although we did discover that variability in MT quality did not have a significant impact on classification accuracy, we would still be interested in seeing whether improved MT quality provided by the latest Microsoft neural network MT systems impacts the classification of customer feedback.

## 7 Acknowledgements

This research is supported by the ADAPT Centre for Digital Content Technology, funded under the Science Foundation Ireland (SFI) Research Centres Programme (Grant 13/RC/2106).

## References

Afli, Haithem, Sorcha McGuire, and Andy Way. Sentiment Translation for low-resourced languages: Experiments on Irish General Election Tweets. In *Proceedings of the 18th International Conference on Intelligent Text Processing and Computational Linguistics*, Budapest, Hungary, 2017.

Bentley, Michael and Batra, Soumya. Giving Voice to Office Customers: Best Practices in How Office Handles Verbatim Text Feedback. In *IEEE International Conference on Big Data*. pp. 3826–3832, 2016.

Burns, Michelle. (2016, February). Kampyle Introduces the NebulaCX Experience Optimizer. Retrieved from http://www.kampyle.com/kampyle-introduces-the-nebulacx-experience-optimizer/

Equiniti. (2017, April). Complaints Management. Retrieved from https://www.equiniticharter.com/services/complaints-management/#.WOH5X2_yt0w

Freshdesk Inc. (2017, February). Creating and sending the Satisfaction Survey. Retrieved from https://support.freshdesk.com/support/solutions/articles/37886-creating-and-sending-the-satisfaction-survey

Inmoment. (2017, April). Software to Improve and Optimize the Customer Experience. Retrieved from http://www.inmoment.com/products/

Keatext Inc. (2016, September). Text Analytics Made Easy. Retrieved from http://www.keatext.ai/

Potharaju, Rahul, Navendu Jain and Cristina Nita-Rotaru. Juggling the Jigsaw: Towards Automated Problem Inference from Network Trouble Tickets. In *10th USENIX Symposium on Network Systems Design and Implementation (NSDI 13)*. pp. 127–141, 2013.

Salameh, Mohammad, Saif M Mohammad, and Svetlana Kiritchenko. Sentiment after translation: A case-study on Arabic social media posts. In *Proceedings of the 2015 Annual Conference of the North American Chapter of the ACL*, pp. 767–777, 2015.

SurveyMonkey Inc. (2017, April). Customer Service and Satisfaction Survey. Retrieved from https://www.surveymonkey.com/r/BHM_Survey

UseResponse. (2017, April). Customer Service & Customer Support are best when automated. Retrieved from https://www.useresponse.com/

Yin, Dawei, Yuening Hu, Jiliang Tang, Tim Daly, Mianwei Zhou, Hua Ouyang, Jianhui Chen, Changsung Kang, Hongbo Deng, Chikashi Nobata, Jean-Mark Langlois, and Yi Chang. Ranking relevance in yahoo search. In *Proceedings of the 22nd ACM SIGKDD International Conference on Knowledge Discovery and Data Mining*. pp. 323–332, 2016. ACM.